\documentclass{ieeeaccess}
\usepackage{cite}
\usepackage{amsmath,amssymb,amsfonts}
\usepackage{algorithmic}
\usepackage{graphicx}
\usepackage{textcomp}
\usepackage{booktabs}
\usepackage{multirow}
\usepackage{graphics}
\usepackage{pifont}
\usepackage[colorlinks,
            linkcolor=blue,       
            anchorcolor=blue,  
            citecolor=blue,        
            ]{hyperref}

\def\BibTeX{{\rm B\kern-.05em{\sc i\kern-.025em b}\kern-.08em
    T\kern-.1667em\lower.7ex\hbox{E}\kern-.125emX}}
\begin{document}
\history{Date of publication xxxx 00, 0000, date of current version xxxx 00, 0000.}
\doi{10.1109/ACCESS.2023.0322000}

\title{AMANet: Advancing SAR Ship Detection with  Adaptive Multi-Hierarchical Attention Network}
\author{\uppercase{Xiaolin Ma}\authorrefmark{1},
\uppercase{Junkai Cheng}\authorrefmark{2}, \uppercase{Aihua Li*}\authorrefmark{1}, \uppercase{Yuhua Zhang}\authorrefmark{1}, and \uppercase{Zhilong Lin*}\authorrefmark{1},
}

\address[1]{Shijiazhuang,Campus,Army Engineering University of PLA, Shijiazhuang, 050003,China (e-mail: xiaolin.ma@163.com)}
\address[2]{School of Automation,Northwestern Polytechnical University, Xian,
710129,China (e-mail: author@lamar.colostate.edu)}

\tfootnote{This work was supported in part by the National Natural Science Foundation of China under Grant 62171467.}

\markboth
{X. Ma \headeretal: AMANet: Advancing SAR Ship Detection with  Adaptive Multi-Hierarchical Attention Network}
{X. Ma \headeretal: AMANet: Advancing SAR Ship Detection with  Adaptive Multi-Hierarchical Attention Network}

\corresp{Corresponding author: Zhilong Lin (e-mail: 15639150607@163.com) and Aihua Li (e-mail: yuandianqi@163.com).}

\begin{abstract}
Recently, methods based on deep learning have been successfully applied to ship detection for synthetic aperture radar (SAR) images. Despite the development of numerous ship detection methodologies, detecting small and coastal ships remains a significant challenge due to the limited features and clutter in coastal environments.
For that, a novel adaptive multi-hierarchical attention module (AMAM) is proposed to learn multi-scale features and adaptively aggregate salient features from various feature layers, even in complex environments.
Specifically, we first fuse information from adjacent feature layers to enhance the detection of smaller targets, thereby achieving multi-scale feature enhancement. 
Then, to filter out the adverse effects of complex backgrounds, we dissect the previously fused multi-level features on the channel, individually excavate the salient regions, and adaptively amalgamate features originating from different channels.
Thirdly, we present a novel adaptive multi-hierarchical attention network (AMANet) by embedding the AMAM between the backbone network and the feature pyramid network (FPN). Besides, the AMAM can be readily inserted between different frameworks to improve object detection.
Lastly, extensive experiments on two large-scale SAR ship detection datasets demonstrate that our AMANet method is superior to state-of-the-art methods.
\end{abstract}


\titlepgskip=-21pt

\maketitle

\section{Introduction}
\label{sec:introduction}

\PARstart{S}{ynthetic} aperture radar (SAR) \cite{bhattacharjee2023deep, gao2022improved, han2020multi} provides high-resolution imaging capabilities that remain unaffected by daylight, weather conditions, and other environmental factors. This makes SAR an indispensable tool for remote sensing applications. Ship detection in SAR images plays a critical role in various domains such as national defense, maritime management, identification of illicit activities, marine transport monitoring, and coastal security enhancement. However, this task presents significant challenges due to sea clutter, ship size variability, and land clutter interference. Consequently, further research is urgently needed to enhance the accuracy of offshore vessel detection in SAR images. This research area is both significant and complex, offering substantial practical implications.

\begin{figure}
    \centering
    \includegraphics[width=1\linewidth]{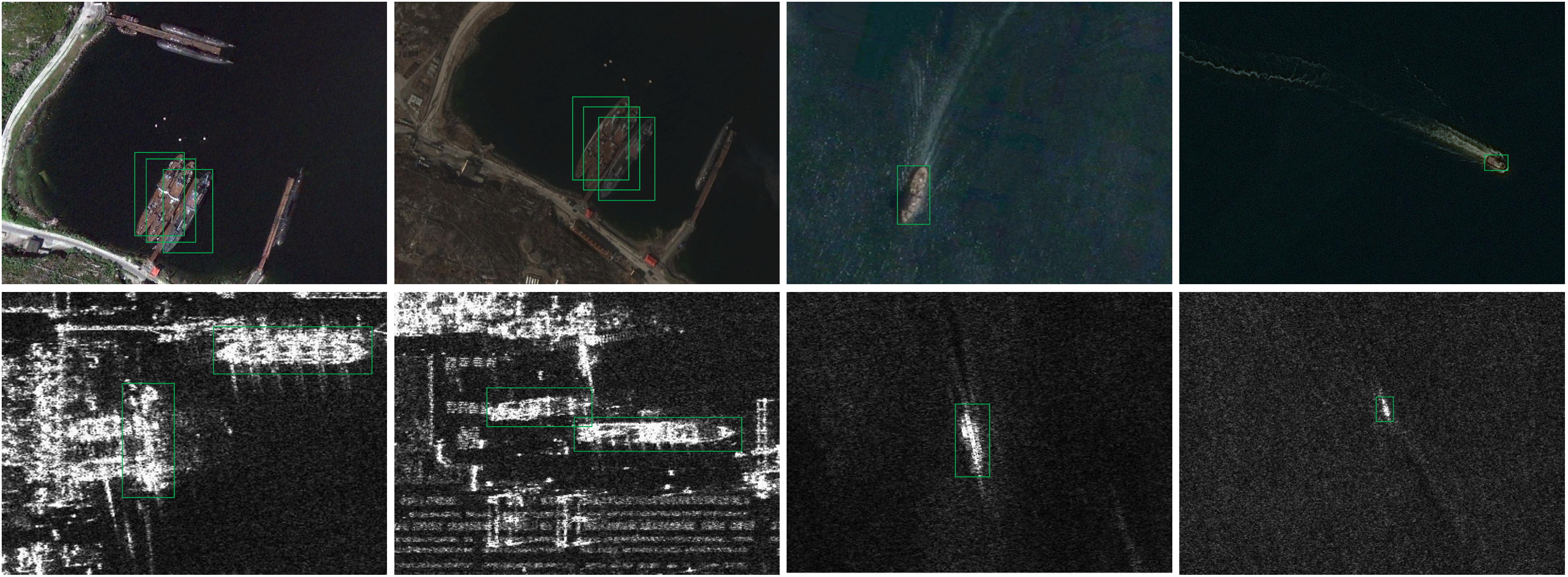}
    \caption{The difference between visible and SAR images. The first row shows visible images, and the second row shows SAR images. The green rectangles enclose the ground truth.}
    \label{fig: vs}
\end{figure}

Convolutional neural networks (CNNs) have been extensively employed in visible image object detection \cite{liu2019large}, delivering remarkable results \cite{fan2019ship}. When applied to ship detection in SAR images, these CNN algorithms have proven highly effective \cite{liu2016ssd, ren2015faster}.
Subsequently, the FPN \cite{lin2017feature} has emerged as a standard solution for detecting ships in multi-scale SAR images. Building on the foundation of FPN, later research has concentrated on Bi-directional FPN to enhance the representation of hierarchical features \cite{liu2018path, tan2020efficientdet}.
However, these methods require further refinement and enhancement to effectively handle extreme-scale changes or scenarios with fewer ship features. As illustrated in Figure \ref{fig: vs}, the top row presents visible images, while the bottom row features SAR images.
SAR images have the distinct advantage of increased sensitivity to metallic objects, significantly aiding ship object detection. 
However, they offer less color texture and other details when compared to visible images, presenting a unique set of challenges.

The attention mechanism has gained significant traction in the field of computer vision. There are three commonly utilized attention methods: spatial attention, channel attention, and combined spatial and channel attention.
Spatial attention methods \cite{jaderberg2015spatial, hu2018gather} generate attention masks across spatial domains, which are employed to select crucial spatial regions or directly predict the most relevant spatial positions. Channel attention methods \cite{hu2018squeeze, wang2020eca}, on the other hand, generate attention masks across the channel domain, which are used to select essential channels.
Methods that combine spatial and channel attention \cite{woo2018cbam, park2018bam} compute temporal and spatial attention masks separately or produce a joint spatiotemporal attention mask to focus on informative regions. However, these attention methods have shown limited improvement in SAR images, which typically have fewer color and texture features. This limitation is particularly evident in ground clutter near the coast, significantly impacting object detection.
As illustrated in Figure \ref{fig: vs}, there is a high similarity between ground clutter and ships in near-shore scenes. Unlike visible images, ships in SAR images cannot be distinguished through color and other features, presenting a unique challenge for detection algorithms. Further, detecting small and coastal ships in coastal environments with limited features and clutter is difficult.

In order to meet the above challenges, we propose a novel AMAM designed to learn multi-scale features and adaptively aggregate salient features from various feature layers, even in complex environments. Our method involves several key steps.
First, we fuse information from adjacent feature layers to enhance the detection of smaller targets, achieving multi-scale feature enhancement. Next, to mitigate the adverse effects of complex backgrounds, we dissect the previously fused multi-level features on the channel, individually excavate salient regions, and adaptively amalgamate features from different channels.
Subsequently, we introduce a novel AMANet by embedding the AMAM between the backbone network and the FPN. The AMAM can be readily inserted between different frameworks to improve object detection.
Finally, extensive experiments on two large-scale SAR ship detection datasets demonstrate the superiority of our AMANet method compared to the state-of-the-art method, highlighting its potential for advancing ship detection in challenging environments.
The main contributions of this article are as follows:

\begin{itemize}
\item[$\bullet$] 
This paper presents a plug-and-play AMAM to learn multi-scale features and adaptively aggregate salient features from various feature layers.

\item[$\bullet$] We propose a novel AMANet to insert AMAM between different frameworks to improve object detection.

\item[$\bullet$] 
We conduct extensive experiments on two large-scale object datasets, demonstrating promising performance gains achieved by AMANet. 
Additionally, numerous ablation studies validate the effectiveness of the core mechanisms
in AMANet for SAR ship detection. 
\end{itemize}

The rest of this paper is organized as follows. Section~\ref{rw}
introduces the related work. Section \ref{method} elaborates our method.
Section \ref{exp} presents the experimental results to show our
method’s superiority. Section \ref{conclusion} concludes this paper.

\section{RELATED WORK}\label{rw}
In this section, we briefly review the most related works of SAR ship detection, multi-scale feature fusion, and attention mechanism.

\subsection{SAR ship detection}
Recently, SAR ship detection \cite{weng2023novel, qiao2022novel}  has gained significant attention in the remote sensing community \cite{zhao2014ship}.
Traditional ship detection methods often rely on techniques like CFAR \cite{farina1986review} or hand-crafted features. 
However, these methods need help in effectively detecting ships across multiple scales.
In recent years, deep learning methods, particularly CNNs, have emerged as a promising solution for ship detection due to their powerful feature representation capabilities. 
CNN-based object detectors can be broadly classified into two categories: two-stage detectors and one-stage detectors~\cite{yasir2023ship}. 
Two-stage detectors \cite{hu2023bag} initially generate candidate regions in the first stage and subsequently classify, identify, and position based on these candidate regions in the second stage.
While these methods often achieve higher detection accuracy, they require more computational resources. 
On the other hand, one-stage detectors, such as SSD, RetinaNet, and YOLO series\cite{redmon2016you, redmon2017yolo9000, redmon2018yolov3, liu2019large, bochkovskiy2020yolov4}, directly predict the category and position coordinates of targets in a single step, eliminating the need for explicit region proposal generation. 
For example, CFIL \cite{weng2023novel} proposes a frequency-domain feature extraction module and feature interaction in the frequency domain to enhance salient features.
MFC \cite{qiao2022novel} proposes a frequency-domain filtering module to achieve dense target feature enhancement.

\subsection{Multi-Scale Feature Fusion}

Multi-scale feature fusion \cite{hassan2020learning, lin2017feature} is essential for object detection by aggregating and enhancing information in SAR ship detection. 
Different methods, such as simple feature fusion, feature pyramid fusion, and cross-scale feature fusion, have been proposed for multi-scale feature fusion \cite{lin2017feature, hu2023robust, yu2018bisenet}. 
Simple feature fusion combines feature maps from adjacent layers to compensate for information loss \cite{wu2022sample} during transmission and consider contextual information\cite{zhao2018psanet, shen2023triplet}. 
For example,  EMRN \cite{shen2021efficient} proposes a multi-resolution features dimension uniform module to fix dimensional features from images of varying resolutions.
DAL \cite{ming2021dynamic} proposes a dynamic anchor learning method, which utilizes the newly defined matching degree to evaluate the localization potential of the anchors comprehensively and carries out a more efficient label assignment process.
The HPGN \cite{shen2021exploring} introduces a pyramidal graph network extract fine-grained image features.
These methods enhance the overall representation of features by incorporating adjacency information.
However, challenges arise when dealing with extreme scale differences in SAR ship detection, leading to compression or blurring of features and information loss.

\begin{figure*}
    \centering
    \includegraphics[width=0.95\textwidth]{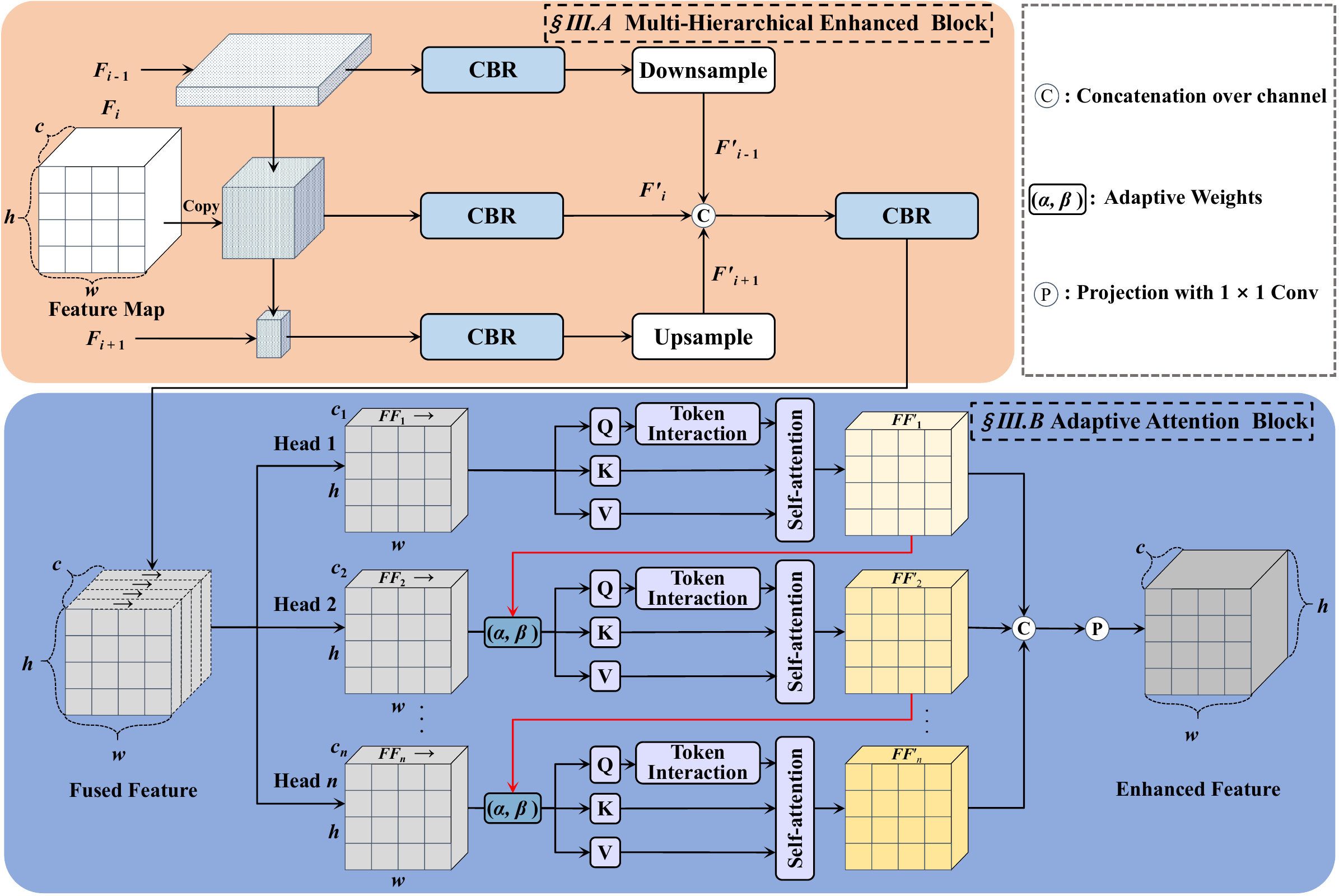}
    \caption{The structure of the AMAM. It consists of two main components: the multi-hierarchical enhanced block (ME) and the adaptive attention block (AA). The ME block leverages the contextual features from adjacent and deeper layers, aiding in accurate ship detection. The AA block splits the fused feature to each attention head, enhancing the diversity of attention maps and allowing for more discrimination to inshore clutter. Note that CBR is Convolution, Batch Normalization, and ReLU. $F_{i}$ is the feature map of the current layer. $c$, $h$, and $w$ are the Fused Feature's channel, height, and width, respectively, and $c_1$ = $c_i$ = $c_n$. $\alpha$, $\beta$ are learnable coefficients.}
    \label{fig: amam}
\end{figure*}

\subsection{Attention Mechanism}

Recently, attention mechanisms \cite{jaderberg2015spatial, shen2022competitive, dai2017deformable, woo2018cbam}  have been gaining increasing attention in computer vision.
For example, 
STN \cite{jaderberg2015spatial} predicts affine transformations to selectively attend to crucial regions in the input. This stage was characterized by a focus on discriminative input features, with DCNs \cite{dai2017deformable} being a notable example.
HSGM \cite{shen2022hsgm, shen2022hsgnet} proposes a hierarchical similarity graph module to relieve the conflict of backbone networks and mine the discriminative features.
SENet \cite{hu2018squeeze} proposes a channel-attention mechanism that implicitly and adaptively predicts essential features. 
CAM proposes a contrastive attention module to enhance local features through many-to-one learning.
GiT \cite{shen2023git}  proposes a structure where graphs and transformers interact constantly, enabling close collaboration between global and local features for vehicle re-identification.
Works such as EMANet \cite{li2019expectation}, CCNet \cite{huang2019ccnet}, and Stand-Alone Networks \cite{ramachandran2019stand} have leveraged self-attention to improve speed, result quality, and generalization capabilities.
Besides, there are also some self-attention and cross-attention mechanisms. For example, PBSL \cite{hen2023pbsl} introduces a co-interaction attention module to highlight relevant features and suppress irrelevant information.
However, these attention methods cannot effectively distinguish ground clutter, which is similar to ships.

\section{Proposed Method} \label{method}

\subsection{Multi-hierarchical enhanced block}
The AMAM introduces a crucial component called the ME block, vital in combining high-level semantic features from deep layers with shallower layers in both top-down and bottom-up directions. The ME block aims to balance preserving important features for accurate predictions and minimizing computationally expensive operations like convolution, pooling, and addition.

As shown in Figure  \ref{fig: amam}, given a feature map $F_{i}$ of size C $\times$ H $\times$ W, where C, H, and W represent the channel, height, and width of the feature diagram.
The module takes three features extracted from the backbone: deep feature $F_{i+1}$ (2C, H/2, W/2), current feature $F_{i}$, and shallow feature $F_{i-1}$ (C/2, 2H, 2W). 
Firstly, the feature processing pipeline starts with convolution, batch normalization, and ReLU (CBR) operations. These operations enhance the features' representational power.
Next, the upsampled deep and downsampled shallow features are incorporated into the current feature through combination. This fusion of features at different scales captures information from larger and smaller contexts.
Finally, the concatenated features undergo further CBR operations, refining and consolidating the information from multiple scales. These operations generate the final fused feature, representing a comprehensive and enriched input data representation.
This fused feature represents the enhanced multi-scale representation of the image. This operation can be formulated as follows:

\begin{equation}
F_{i}^{'}  = \operatorname{CBR}(F_{i}),
\end{equation}
where $F_{i}$ (C, H, W) represents the current feature. $\operatorname{CBR}$ represents the convolution, batch normalization, and ReLU operations. $F_{i}^{'}$ (C, H, W) denotes the current features after unified dimension processing.

\begin{equation}
F_{i-1}^{'}  = \operatorname{Upsample}(\operatorname{CBR}(F_{i-1})),
\end{equation}
where $F_{i-1}$ (C/2, 2H, 2W) represents the shallow feature. $\operatorname{Upsamle}$  refer to the upsampling  operations. $F_{i-1}^{'}$ (C, H, W) denotes the shallow features after unified dimension processing.

\begin{equation}
F_{i+1}^{'}  = \operatorname{Downsample}(\operatorname{CBR}(F_{i+1})),
\end{equation}
where $F_{i+1}$ (2C, H/2, W/2) represents the deep feature. $\operatorname{Downsample}$ refer to the downsampling operations. $F_{i+1}^{'}$ (C, H, W) denotes the deep features after unified dimension processing.

\begin{equation}
FF = \operatorname{CBR}(Concat[F_{i-1}^{'}, F_{i}^{'}, F_{i+1}^{'})]),
\end{equation}
where $FF$ (C, H, W) denotes the fused feature. $Concat$  means channel concatenation. The ME block employs concatenation and reorganization operations to fuse features. What sets the ME block apart from existing concatenation methods used in state-of-the-art techniques is its incorporation of contextual features from adjacent and deeper layers. This means that the ME block not only fuses features from the current scale but also leverages features from three adjacent scales (shallow, current, deep) of the backbone network. By doing so, it enriches the features and enhances the detection performance.

The ME block in AMAM improves accuracy and efficiency, as it effectively captures multi-scale information and integrates it into the feature representation process. By leveraging the contextual features from adjacent and deeper layers, the ME block enables the model to extract more comprehensive and discriminative features, aiding in accurate ship detection, particularly for small and coastal ships in complex coastal environments.

\begin{figure*}
    \centering
    \includegraphics[width=0.95\textwidth]{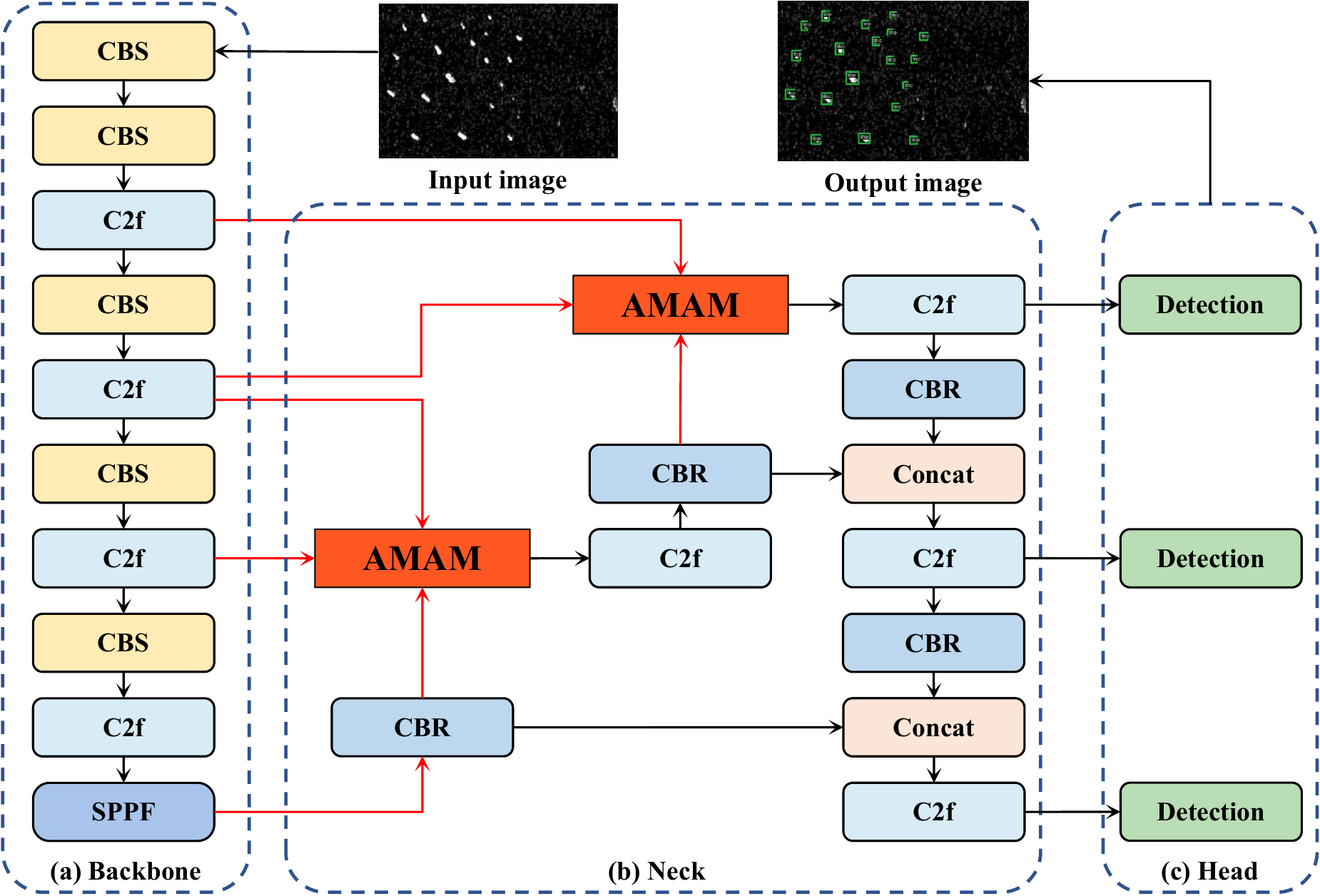}
    \caption{The network structure of the proposed AMANet. The figure showcases the integration of AMAM into the YOLO model (based on YOLOv8s), requiring additional backbone network features. CBS represents convolution, batch normalization, and SiLU activation. SPPF denotes the spatial pyramid pooling fusion module. The C2F module is a lightweight module inspired by c3 and incorporates ideas from ELAN.}
    \label{fig: overall}
\end{figure*}

\begin{table*}[t]
\caption{\textbf{Comparison (\%) between SAR ship detection methods on the SSDD dataset.} The "-" symbol indicates that the corresponding paper did not report the results.}
\centering
\renewcommand{\arraystretch}{1.5}
\begin{tabular}{l||cccc|c}
\hline
Method                                         & $AP_{0.5:0.95}$                 & $AP_{0.5}$                      & Precision (IoU=0.5)             & Recall (IoU=0.5)                & Reference                        \\ \hline
ImYOLOv4\cite{gao2022improved} & -                               & 94.16                           & 93.54                           & 90.95                           & 2022 IEEE Access                   \\
PPA-Net\cite{tang2023ppa}     & -                               & 95.19                           & 95.22                           & 91.22                           & 2023 Remote Sensing                   \\
A-BFPN\cite{li2022bfpn}       & 59.60                           & 96.80                           & -                               & -                               & 2022 Remote Sensing                   \\
FEPS-Net\cite{bai2023feature} & 59.90                           & 96.00                           & -                               & -                               & 2023 IEEE JSTAEORS                    \\
HR-SDNet\cite{wei2020precise} & 64.60                               & 97.90                           & -                           & -                           & 2020 Remote Sensing                   \\
SSE-Ship\cite{zheng2023sse}   & 64.70                           & 96.40                           & 94.40                           & 94.00                           & 2023 OJAS \\
CS$^{n}$Net\cite{chen2023cs}  & 64.90                           & 97.10                           & -                               & -                               & 2023 IEEE TGRS                        \\
LssDet \cite{yan2022lssdet}  & 68.10                           & 96.70 & -                           & -                           & 2022 Remote Sensing                      \\ \hline
AMANet                                       & \textbf{74.20} & \textbf{98.50}                           & \textbf{97.47} & \textbf{96.60} & Ours                                \\ \hline
\end{tabular}

\vspace{0.5mm}
\label{table 1}
    \vspace{-3mm}
\end{table*}

\begin{table}[t]
\caption{\textbf{Comparison (\%) on the HRSID dataset.} The "-" symbol indicates that the corresponding paper did not report the results.}
    \centering
    \renewcommand{\arraystretch}{1.5}
    \begin{tabular}{l||cc|c}
\hline
Method & $AP_{0.5:0.95}$ & $AP_{0.5}$ & Reference \\ \hline
CSD-YOLO\cite{chen2023multi}      & -               & 86.10      & 2023 Remote Sensing         \\
Quad-FPN\cite{zhang2021quad}      & -               & 86.12      & 2021 Remote Sensing         \\
MEA-Net\cite{guo2022mea}      & -               & 89.06      & 2022 Remote Sensing       \\
PPA-Net\cite{tang2023ppa}      & -               & 89.27      & 2023 Remote Sensing         \\
CS$^{n}$Net\cite{chen2023cs}      & -               & 91.20      & 2023 IEEE TGRS         \\
FINet\cite{hu2022finet}      & -           & 90.50      & 2022 IEEE TGRS          \\
Improved PRDet\cite{chen2023cs}      & 59.80           & 90.70      & 2023 IEEE TGRS         \\
DSDet\cite{sun2021dsdet}      & 60.50           & 90.70      & 2021 Remote Sensing         \\
CenterNet2\cite{sun2022scan}      & 64.50           & 89.50      & 2022 IEEE TGRS          \\
SRDet\cite{lv2023anchor}      & 66.10           & 90.60      & 2023 Remote Sensing         \\ 
\hline
AMANet     & \textbf{68.90}               & \textbf{91.40}         & Ours        \\ \hline
\end{tabular}
    \vspace{0.5mm}
\label{table 2}
    \vspace{-3mm}
\end{table}

\subsection{Adaptive attention block}

In the context of multi-head self-attention, one of the significant challenges is the redundancy present in attention heads, which can lead to computational inefficiency. We took inspiration from cascaded group attention to overcome this issue and developed an efficient AA block.
The AA block addresses the problem by introducing different splits of the full features to each attention head, enabling an explicit decomposition of attention computation across the heads. Furthermore, the Q, K, and V layers learn projections on features with richer information. We achieve computational efficiency and reduce computation overhead by utilizing feature splits instead of the full features for each head.
To aggregate information from different heads, the AA block adds the output of each head to the subsequent head, progressively refining the feature representations. This iterative aggregation process helps enhance the diversity of attention maps by introducing distinct feature splits to each attention head.
Additionally, the concatenation of attention heads increases the network depth, enhancing the model's capacity. Importantly, this increase in depth comes with only a marginal rise in latency overhead, as the attention map computation within each head utilizes smaller QK channel dimensions. 
This attention aggregation can be formulated as follows:

\begin{equation}
\widetilde{FF}_{i}=\operatorname{\textit{Selfattn}}\left(FF_{i}^{\prime} W_{i}^{\mathrm{Q}}, FF_{i}^{\prime} W_{i}^{\mathrm{K}}, FF_{i}^{\prime} W_{i}^{\mathrm{V}}\right),
\end{equation}
where, $\widetilde{FF}{i}$ (C/$n$, H, W) and $FF{i}^{\prime}$ (C/$n$, H, W) represent the input and output of the $i$-th head, respectively. $W_{i}^{Q}$, $W_{i}^{K}$, and $W_{i}^{V}$ are projection layers that map the input feature split into different subspaces.

We initialize the first output head as the same as the first input head:

\begin{equation}
FF_{1}^{\prime} = FF_{1},
\end{equation}
where $FF_{i}$ (C/$n$, H, W) represents the $i$-th split of the input feature $FF$ (C, H, W), i.e., $FF = [FF_{1}, FF_{2}, \ldots, FF_{h}]$, and $1 \leq i \leq h$.

The subsequent output heads are obtained by aggregating the previous output head $\widetilde{FF}{i}$ and the current input head $F{i+1}$:

\begin{equation}
FF_{i+1}^{\prime} = \alpha \cdot \widetilde{FF_i} + \beta \cdot FF_{i+1}, \quad 1 \leq i \leq h, \quad \alpha + \beta = 1,
\end{equation}
here, $\alpha$ and $\beta$ are learnable parameters that adaptively adjust the weight coefficients of $\widetilde{FF}_{i}$ and $\textit{FF}_{i+1}$ to improve information aggregation between different heads.

Finally, we concatenate the output heads $\widetilde{FF}_{1}, \widetilde{FF}_{2}, \ldots, \widetilde{FF}_{h}$ and project them back to the dimension consistent with the input:

\begin{equation}
\widetilde{FF}=\operatorname{\textit{Concat}}\left[\widetilde{FF_1}, \widetilde{FF_2}, \ldots, \widetilde{FF_h}\right]_{i=1: h} W^{\mathrm{P}},
\end{equation}
here, $\widetilde{FF}$ is the enhanced feature, $h$ is the total number of heads, $FF_{i}^{\prime}$ represents the input of the $i$-th head's self-attention, and $W^{\mathrm{P}}$ is a linear layer that projects the concatenated output features back to the dimension consistent with the input.

Incorporating the AA block in our proposed AMANet brings two notable advantages. Firstly, introducing distinct feature splits to each attention head enhances the diversity of attention maps, leading to improved feature representation. Secondly, concatenating attention heads increases the model capacity, allowing for more expressive power in capturing complex relationships within the data. These benefits are achieved with minimal additional computational cost, making the AA block an efficient and effective component of the AMANet architecture.

\subsection{OVERALL FRAMEWORK}

The proposed method's overall scheme and the network architecture of AMANet based on YOLOv8s are depicted in Figure \ref{fig: overall}. 
The YOLOv8s' architecture comprises three main modules: the convolution, batch normalization, SiLU activation (CBS) module, the CBR module, the spatial pyramid pooling fusion (SPPF) module, and the C2F (c3-inspired lightweight module with ideas from ELAN) module.

Firstly, the CBS module consists of a 3 × 3 convolutional layer, a Batch Normalization layer, and a SiLU activation function.
This configuration enables the selection of models with high efficiency and accuracy. 
The CBS module mitigates the risk of gradient dispersion by reusing features and retaining most of the original information.
This results in effective feature representation and preservation.
Secondly, inspired by the spatial pyramid pooling (SPP) structure from SPPNet, the SPPF module enhances classification accuracy by extracting and fusing high-level features. 
It employs multiple maximum pooling operations during the fusion process to capture a wide range of high-level semantic features. 
This allows the network to effectively incorporate contextual information and improve the discriminative power of the features.
Thirdly, the C2F module builds upon the C3 module and incorporates ideas from the efficient, lightweight, anchor-free network (ELAN).
It provides a fast and efficient implementation while achieving optimal performance. 
This module plays a crucial role in the network architecture, enabling efficient feature extraction and representation.

\section{EXPERIMENT AND ANALYSIS} \label{exp}
\label{sec:guidelines}

\subsection{DATASETS}
To validate superiority, the proposed AMANet is compared with multiple state-of-the-art methods on SSDD and HRSID datasets.  
\subsubsection{SSDD}
The SSDD~\cite{li2017ship, zhang2021sar} dataset encompasses diverse ship types without specific constraints. 
It primarily comprises data captured in HH, HV, VV, and VH polarization modes. 
Comprising 1160 images, each encapsulates 2456 ships of varying dimensions and quantities.
Following~\cite{zhang2021sar}, it is divided into 928 images for training and 232 images for testing, and it is worth mentioning that the test set includes 46 inshore images and 186 offshore images.

\subsubsection{HRSID}
The HRSID~\cite{wei2020hrsid} dataset contains 5604 high-resolution SAR images and 16,951 ship instances. The HRSID dataset includes SAR images with different resolutions, polarizations, sea states, sea areas, and coastal ports. Following~\cite{wei2020hrsid}, it is divided into 65\% for training and 35\% for testing. 

\begin{table}[t]
\caption{\textbf{Test Result (\%) of Different models on inshore and offshore data in the SSDD dataset.}}
\centering
\renewcommand{\arraystretch}{1.5}
\begin{tabular}{l||c|c|c}
\hline
\multirow{2}{*}{Method}                        & Inshore                         & Offshore                        & \multirow{2}{*}{Reference} \\ \cline{2-3}
                                               & $AP_{0.5:0.95}$                 & $AP_{0.5:0.95}$                 &                            \\ \hline
Swin-PAFF\cite{zhang2023swin}                                      & 37.00                           & 60.30                           & 2023 CMC                   \\
FEPS-Net\cite{bai2023feature} & 47.10                           & 64.50                           & 2023 IEEE JSTAEORS         \\
CS$^{n}$Net\cite{chen2023cs}  & 53.10                           & 64.60                           & 2023 IEEE TGRS             \\
SW-Net\cite{qu2023sw}                                         & 53.50                           & 59.69                           & 2023 SIVP                  \\ \hline
AMANet                                         & \textbf{68.80} & \textbf{76.30} & Ours                          \\ \hline
\end{tabular}
\vspace{0.5mm}
\label{table 3}
    \vspace{-3mm}
\end{table}

\subsection{Evaluation Metrics}
Following \cite{li2017ship, zhang2021sar, wei2020hrsid}, the evaluation metrics used to select the optimal model for maritime remote sensing targets are precision ($P$), recall ($R$), and average precision ($AP$).
$P$ is calculated as the ratio of true positive detections to the total number of positive detections, and it measures the model's accuracy in identifying relevant targets. The formula for $P$ is given by:

\begin{equation}
P = \frac{TP}{TP + FP},
\end{equation}
$TP$ represents the number of true positive detections, and $FP$ represents the number of false positive detections.

Recall ($R$), also known as the true positive rate or sensitivity, is calculated as the ratio of true positive detections to the total number of ground truth positive samples.
Recall measures the ability of the model to identify all relevant targets correctly. The formula for recall is given by:

\begin{equation}
R = \frac{TP}{TP + FN},
\end{equation}
where $FN$ represents the number of false negative detections.

$AP$ is a commonly used metric in object detection tasks. 
The formula for calculating $AP$ involves the precision-recall curve and the area under the curve. 
The formula for $AP$ is given by:

\begin{equation}
AP=\int_0^1 P(R) d R.
\end{equation}

\subsection{Implementation Details}\label{formats}
The experiments are based on the Pytorch 1.10.1 framework and are computed using an NVIDIA RTX3090 (with 24GB of video memory) GPU and CUDA11.3 environment. 
We used the YOLOv8s as a baseline, and network improvements are made on this basis.
In the training process, following \cite{shen2021competitive}, we set the momentum parameter to 0.937, the batch size to 16, and trained 500 epochs.
We used a periodic learning rate and a periodic learning rate and Warm-Up method to warm up the learning rate, where the initial learning rate was set to 0.01. 
In the Warm-Up \cite{fu2022bag} phase, the learning rate of each iteration was updated to 0.1 using linear interpolation. 
After that, we updated the learning rate using the cosine annealing algorithm, and finally, the learning rate dropped to 0.002.

\subsection{Comparison with State-of-the-art Methods}
This section presents the results obtained by the proposed AMANet, compared to a baseline model and state-of-the-art SAR ship detection methods using the SSDD and HRSID datasets. 

\subsubsection{Comparisons on SSDD ship}

The experimental findings, highlighting the performance of AMANet in comparison to other models, are summarized in Table \ref{table 1} based on SSDD. 
AMANet exhibits exceptional performance, surpassing other advanced object detection models. It achieved an $AP_{0.5:0.95}$ score of 74.20\% and an impressive $AP_{0.5}$ score of 98.50\% on the SSDD dataset.
Compared to the self-attention and multi-scale methods, CS$^{n}$Net\cite{chen2023cs}, AMANET has shown a notable improvement. It achieved a 9.30\% increase in the $AP_{0.5:0.95}$ index and a 1.40\% increase in the $AP_{0.5}$ index. These results demonstrate that the ME block effectively integrates multi-scale features and accurately localizes ship targets in SAR images.
When compared to the combined attention-based method and multi-scale method LssDet \cite{yan2022lssdet}, AMANET achieved a 6.10\% increase in $AP_{0.5:0.95}$ and a 1.80\% increase in $AP_{0.5}$. 
In summary, the results indicate that the AA Block in AMANET performs better in focusing on ship targets in SAR images.
The superior performance of AMANet can be attributed to its effective integration of the feature fusion technique and the adaptive multi-hierarchical attention method. 
By fusing information from adjacent feature layers, AMANet improves the representation of multi-scale features and enhances the detection of smaller targets.

\subsubsection{Comparisons on HRSID ship}

Similar to the SSDD dataset, we continue to conduct experiments on the HRSID dataset, and the experimental results are as follows.
As shown in Table\ref{table 2}, the AMANet achieve the best result with 68.90\% and 91.40\% on the $AP_{0.5:0.95}$ and $AP_{0.5}$ respectively.
Compared with combined attention-based methods, AMANet significantly surpasses SRDet \cite{lv2023anchor} in $AP_{0.5:0.95}$ and $AP_{0.5}$ metrics, for example, It exceeds 2.80\% on the $AP_{0.5:0.95}$ and exceeds 0.80\% on the $AP_{0.5}$, which shows that AMANet can better pay attention to ships in near-shore ground clutter.
Compared with multi-scale based methods (i.e., PPA-Net \cite{tang2023ppa}), AMANet outperforms it, for example, leading by 2.13\% on the $AP_{0.5}$ metric, which shows that AMANet can be more accurate to locate ship targets.

\subsubsection{Comparisons in inshore and offshore scenes}

The inshore data contains significant background information, introducing interference and false detections. 
On the other hand, the offshore data consists of densely distributed small targets, which can result in missed detection issues. 
To further validate the superior performance of AMANet in complex backgrounds, separate accuracy tests were conducted on the inshore and offshore data. 
The test results are presented in Table \ref{table 3}.

Our model demonstrates anti-interference solid capability, as evidenced by the experimental findings. 
It achieved the highest detection accuracy on the inshore and offshore test sets, with 68.80\% and 76.30\% for $AP_{0.5:0.95}$, respectively.
Compared to the spatial feature enhancement and weight-guided fusion method SW-Net\cite{qu2023sw}, the proposed method in this article achieved significant improvements. Specifically, in both near-shore and offshore scenarios, there was an increase of 15.30\% and 16.61\%, respectively, in the $AP_{0.5:0.95}$ metric.
Similarly, when compared to the self-attention and multi-scale method CS$^{n}$Net\cite{chen2023cs}, the proposed method demonstrated notable enhancements. In the near-shore and offshore scenarios, there was an increase of 15.70\% and 11.70\%, respectively, in the $AP_{0.5:0.95}$ metric. These results indicate that the method presented in this article excels in handling complex scenarios, particularly in detecting near-shore ship targets.

\begin{table}[t]
\caption{\textbf{Ablation experiments (\%) of AMAM on YOLOv8s.}}
\centering
\renewcommand{\arraystretch}{1.5}
\begin{tabular}{c|cc|c|c}
\hline
\multirow{2}{*}{No.} & \multicolumn{2}{c|}{Settings}                                                                                                     & SSDD                            & HRSID                           \\ \cline{2-5} 
                     & ME                                                              & AA                                                              & $AP_{0.5:0.95} $                & $AP_{0.5:0.95} $                \\ \hline
1                    & \textcolor{grey}{\ding{55}} & \textcolor{grey}{\ding{55}} & 72.10                           & 66.20                           \\
2                    & \textcolor{grey}{\ding{55}} & $\checkmark$                                                    & 73.10                           & 66.70                           \\
3                    & $\checkmark$                                                    & \textcolor{grey}{\ding{55}} & 73.30                           & 67.60                           \\
4                    & $\checkmark$                                                    & $\checkmark$                                                    & \textbf{74.20} & \textbf{68.90} \\ \hline
\end{tabular}
    \vspace{0.5mm}
\label{table 4}
    \vspace{-3mm}
\end{table}

\subsection{Ablation Studies and Analysis}
The comparison results presented in Table \ref{table 1}, Table \ref{table 2}, and Table \ref{table 3} demonstrate that the proposed AMANet method is superior to many state-of-the-art SAR ship detection methods.
By combining the ME and AA blocks, the AMAM can effectively overcome the small targets and complex inshore scene background, contributing to the surpassing performance. 
To further verify the effects of ME and AA blocks in AMAM, the proposed AMANet method is comprehensively analyzed from six aspects to investigate the logic behind its superiority.
(1) Role of AMAM.
(2) Influence of number of heads in AA block. 
(3) Comparisons on fusion functions in AA block.
(4) Universality for different YOLO.
(5) Effects of different attention mechanisms
(6) Visualization. 

\subsubsection{Role of AMAM}

To comprehensively analyze the performance improvement of AMAM with the ME and AA blocks, we conducted an ablation experiment consisting of four experimental sets.
The No.1 group data represents the baseline YOLOv8s experiment results.
The No.2 group data illustrates the performance improvement achieved by incorporating the AA block, resulting in a 1.00\% and 0.50\% increase in $AP_{0.5:0.95}$ on the SSDD and HRSID datasets, respectively.
The No.3 group data demonstrates the performance enhancement obtained by introducing the ME block, leading to a 1.20\% and 1.40\% increase in $AP_{0.5:0.95}$ on the SSDD and HRSID datasets, respectively. 
Finally, the No.4 group data represents the experiment results of AMANet on the SSDD and HRSID datasets. Compared to the baseline, the $AP_{0.5:0.95}$ indicators increased by 2.10\% and 2.70\% on the SSDD and HRSID datasets, respectively. 
These results clearly indicate that both the ME and AA blocks have the potential to improve the model's performance individually. Moreover, when combined, they synergistically bring even more significant improvements.

\subsubsection{Influence of number of heads in AA block}

\begin{figure}
    \centering
    \includegraphics[width=0.45\textwidth]{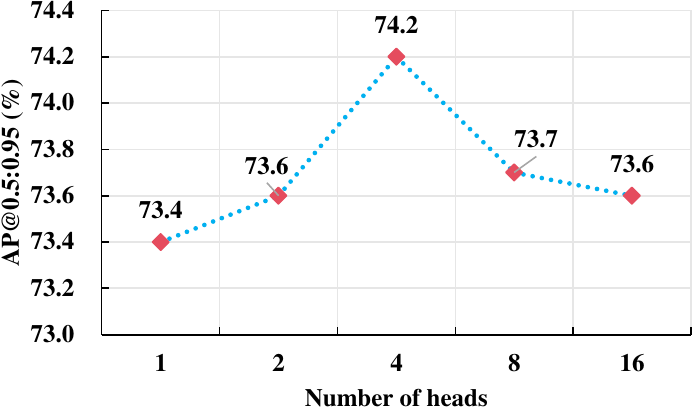}
    \caption{Impact of number of heads in AMAM module on YOLOv8s}
    \label{fig numberofheads}
\end{figure}

We explore the influence of the number of heads in the AA block. 
As shown in Figure \ref{fig numberofheads}, it is evident that the number of attention heads influences the model's performance in the adaptive attention stage. 
Firstly, when using a single attention head, the model achieved an AP of 73.40\%. Secondly, as we increased the number of heads to 2, 4, and 8, the AP scores improved to 73.60\%, 74.20\%, and 73.70\%, respectively. This indicates that employing multiple attention heads can enhance the model's performance, resulting in higher AP scores. Thirdly, we observed a slight decrease in AP when the number of heads increased to 16, reaching a value of 73.60\%. This suggests that there is an optimal range for the number of attention heads, beyond which the performance may start to plateau or decline.

The observed trend suggests that increasing the number of attention heads improves the model's performance, indicating the importance of capturing diverse and discriminative features. 
With more heads, the model can simultaneously attend to different regions of interest, enhancing its ability to capture fine-grained details and subtle variations in the SAR ship images. 
This leads to improved detection accuracy and a higher AP score.
However, when the number of attention heads becomes excessively large, as seen in the case of 16 heads, the performance starts to plateau or even slightly decline.
This may be attributed to the model's increased complexity and potential redundancy in attending to multiple regions with similar characteristics. 
As a result, the model's ability to discriminate between different ship instances may be compromised, leading to a slight decrease in AP.
We selected 4 attention heads for our other experiments based on these results. 
This configuration achieved the highest AP score of 74.20\% among the tested options, striking a balance between capturing diverse features and avoiding redundancy.

\subsubsection{Comparisons on fusion functions in AA block}

\begin{figure}
    \centering
    \includegraphics[width=0.45\textwidth]{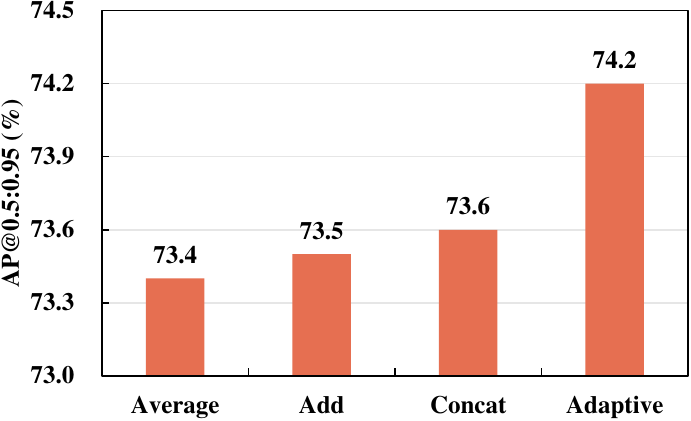}
    \caption{Impact of fusion functions in adaptive attention stage.}
    \label{fig fusion}
\end{figure}

The AMAM incorporates learnable parameters, $\alpha$ and $\beta$, in the adaptive attention block to dynamically adjust the information aggregation between different heads.
To further investigate the impact of alternative fusion methods on the performance of AMANet, we conducted additional experiments.

In Figure \ref{fig fusion}, we compare the performance of AMANet using different fusion functions in the adaptive attention stage. 
The fusion methods evaluated include Average, Add, Concat, and Adaptive (as used in this study).
Firstly, the results reveal that when Average fusion is employed between different heads, the achieved $AP_{0.5:0.95}$ reaches 73.40\%.
Secondly, when Add Fusion is used, the performance slightly improves to 73.50\%.
Thirdly, when Concat fusion is employed, the $AP_{0.5:0.95}$ further increases to 73.60\%. 
Finally, it is noteworthy that when the proposed Adaptive fusion is utilized, the optimal result of 74.20\% is achieved.
These experimental findings emphasize the importance of the adaptive fusion method proposed in this article. 
By dynamically adjusting the information aggregation with the help of learnable parameters, the Adaptive fusion enables AMANet to achieve superior performance, surpassing the alternative fusion methods. 

\subsubsection{Universality for different YOLO}

\begin{figure}
    \centering
    \includegraphics[width=0.45\textwidth]{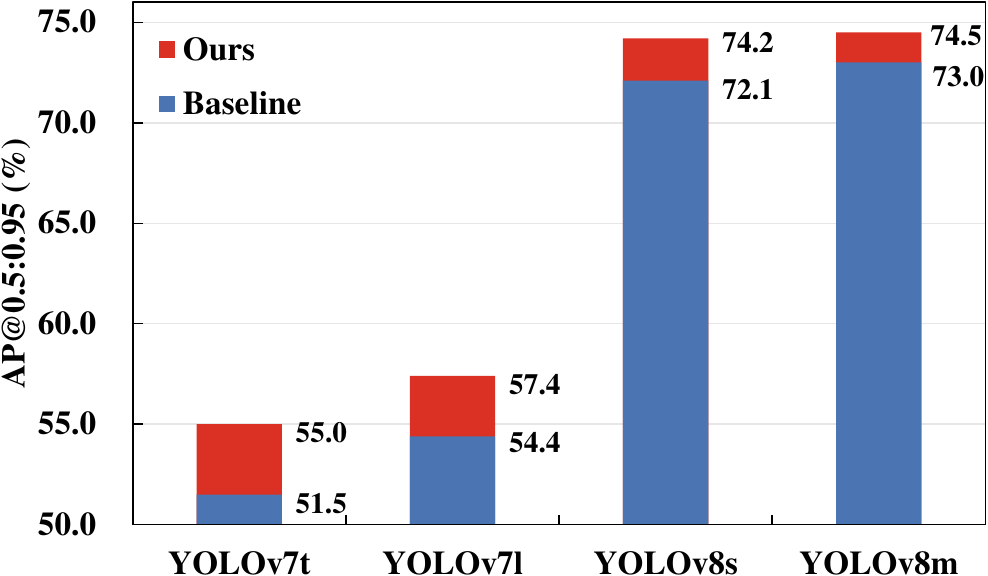}
    \caption{Impact of AMAM module on YOLOv7t, YOLOv7l, YOLOv8s and YOLOv8m models.}
    \label{fig generalization}
\end{figure}

\begin{figure*}
    \centering
    \includegraphics[width=0.95\textwidth]{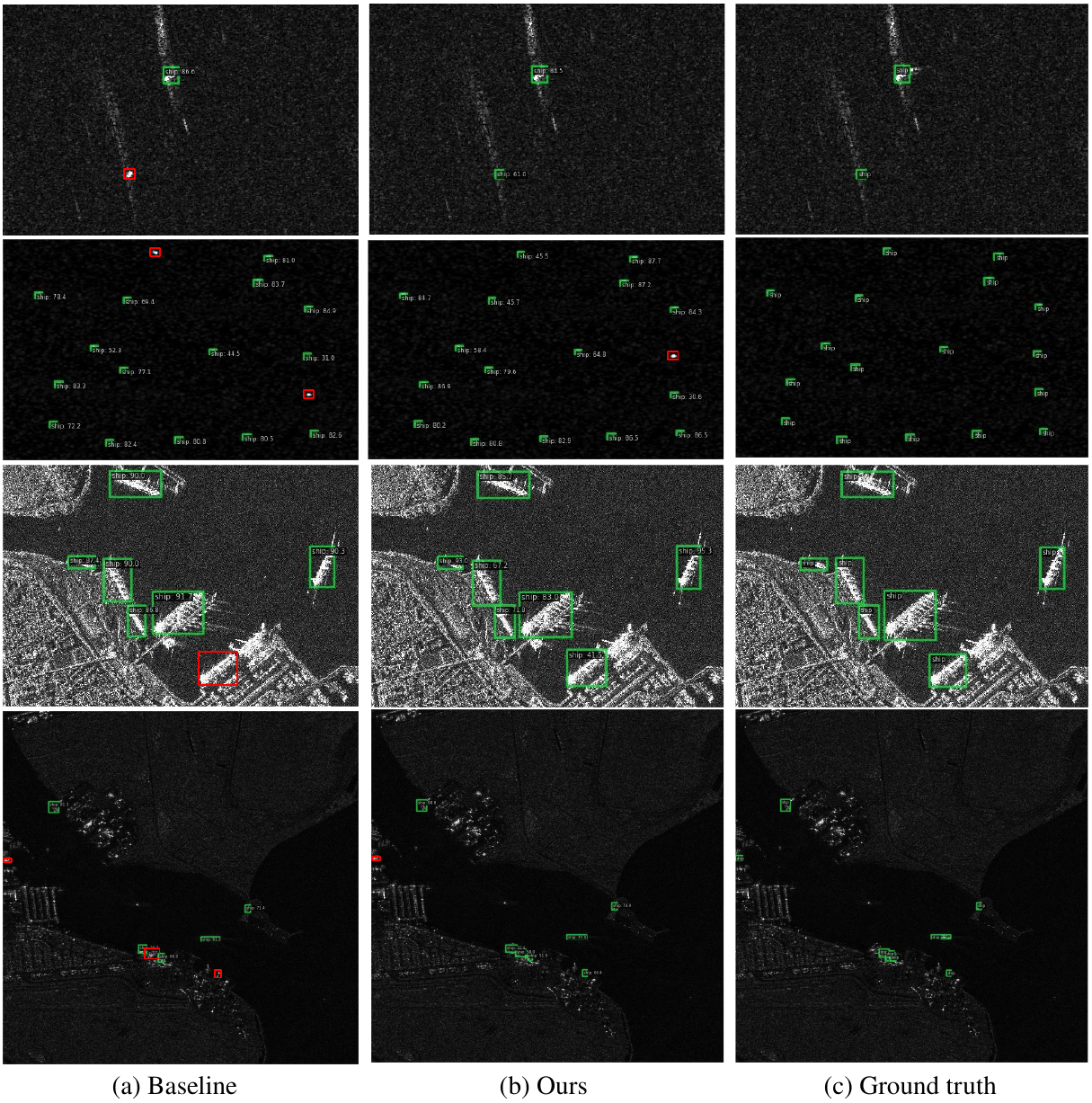}
    \caption{Detection results in SSDD and HRSID datasets with YOLOv8s and AMANet. (a) and (b) show the detection results of the YOLOv8s(baseline) and AMANet; (c) represents the ground truth. The red bounding boxes indicate the prcsene of error detection and omission detection for ground truth.}
    \label{fig result}
\end{figure*}

To evaluate the universality and robustness of the proposed model, we extended the application of the AMAM to different YOLO models, namely YOLOv7t, YOLOv7l, and YOLOv8m. 
The experimental results, as depicted in Figure \ref{fig generalization}, highlight the impact of incorporating the AMAM on the performance of these models.
The results obtained from the evaluation indicate that the inclusion of the AMAM brings about significant improvements in terms of $AP_{0.5:0.95}$ across all the evaluated YOLO models.
Firstly, when considering YOLOv7t, the integration of the AMAM led to a remarkable 3.50\% increase in $AP_{0.5:0.95}$. As a result, the final achieved $AP_{0.5:0.95}$ value for YOLOv7t reached 55.00\%.
Moving on to YOLOv7l, the AMAM delivered a substantial 3.00\% improvement in $AP_{0.5:0.95}$. Consequently, the final achieved $AP_{0.5:0.95}$ value for YOLOv7l reached an impressive 57.40\%.
Furthermore, for YOLOv8s, the inclusion of the AMAM resulted in a noteworthy 2.10\% increase in $AP_{0.5:0.95}$. As a result, the final achieved $AP_{0.5:0.95}$ value for YOLOv8s reached a high of 74.20\%.
Lastly, for YOLOv8m, the AMAM contributed to a notable 1.50\% increase in $AP_{0.5:0.95}$. Consequently, the final achieved $AP_{0.5:0.95}$ value for YOLOv8m reached a commendable 74.50\%.
These findings not only demonstrate the effectiveness of the AMAM in significantly enhancing the performance of YOLOv8s but also highlight its positive impact on other YOLO variants, including YOLOv7t, YOLOv7l, and YOLOv8m. The consistent improvements observed across different model variations further validate the generalization and versatility of the AMAM module.
This highlights its potential as a valuable component for enhancing ship detection performance in various YOLO-based architectures, contributing to the overall universality and applicability of the proposed method.

\subsubsection{Effects of different attention mechanisms}

Further, we compared AMAM with commonly used attention mechanisms such as GE \cite{hu2018gather}, CBAM \cite{woo2018cbam}, and SE \cite{hu2018squeeze}. The results are presented in Table \ref{table 5}.
The model's accuracy improved across the board after incorporating different attention mechanisms. However, when the AMAM was introduced, it led to the most improvement, with a 2.10\% increase compared to $AP_{0.5:0.95}$ to the baseline. On the other hand, introducing the GE, CBAM, and SE attention mechanisms resulted in improvements of 0.50\%, 0.70\%, and 0.80\% in $AP_{0.5:0.95}$, respectively.
In conclusion, the AMAM proves to be highly effective in focusing on ships in SAR images, particularly in the presence of clutter from land and sea. It outperforms other attention mechanisms, showcasing its ability to enhance ship detection performance in challenging scenarios.

\subsubsection{Visualization}

The visualization data presented in Figure \ref{fig result} provides a comprehensive analysis of the detection results, highlighting the performance improvements achieved by AMANet.
The Figure showcases four representative detection examples, each demonstrating the model's effectiveness compared to the baseline YOLOv8s. The first two images depict SAR images containing small ship targets. 
The baseline YOLOv8s results shown in Figure \ref{fig result} (a) exhibit missed detections, as indicated by the red rectangles, where the baseline model fails to detect three small ship targets.
However, the detection results of AMANet in Figure \ref{fig result} (b) miss detect one small ship target, as denoted by the red rectangle.
This demonstrates the superior performance of AMANet in accurately detecting small targets, thereby improving the overall ship detection capability.

\begin{table}[t]
\caption{\textbf{Comparison (\%) of different attention mechanisms.}}
\centering
\renewcommand{\arraystretch}{1.5}
\begin{tabular}{l|c|c}
\hline
Method         & $AP_{0.5:0.95} $ & Type                          \\ \hline
Baseline      & 72.10            & Baseline                      \\
 + GE\cite{hu2018gather}   & 72.60            & Spatial attention             \\
 + CBAM \cite{woo2018cbam} & 72.80            & Spatial \& channel attention \\
 + SE\cite{hu2018squeeze}   & 72.90            & Channel attention             \\ \hline
 +AMAM (Ours)   & 74.20            & Ours                          \\ \hline
\end{tabular}
    \vspace{0.5mm}
\label{table 5}
    \vspace{-3mm}
\end{table}

The latter two images represent SAR images of inshore scenes, which include multiple ship targets. 
Figure \ref{fig result} (a) reveals the limitations of the baseline YOLOv8s, with both missed detections and false alarms.
In contrast, the detection results of AMANet shown in Figure \ref{fig result} (b) are more accurate, with improved precision and recall.
The AMANet has less error detection and omission detection, as demonstrated by the red rectangles.
These visualization examples highlight the superior performance of AMANet in ship detection tasks.
By effectively integrating the ME and AA blocks, AMANet demonstrates improved accuracy and robustness.
It successfully detects small ship targets, accurately identifies ships in complex near-shore scenes, and outperforms the baseline YOLOv8s.
The visualization data provides a clear and visual representation of the model's capabilities, reinforcing the experimental results and demonstrating the practical significance of the proposed AMANet in real-world ship detection scenarios.

\section{Conclusion}\label{conclusion}

In conclusion, this paper presents a novel adaptive multi-hierarchical attention module (AMAM) and network (AMANet) to address the significant challenge of detecting small and coastal ships in SAR images. The AMAM is designed to learn multi-scale features and adaptively aggregate salient features from various feature layers, even in complex environments.
The methodology involves fusing information from adjacent feature layers to enhance the detection of smaller targets, thereby achieving multi-scale feature enhancement. Furthermore, to mitigate the adverse effects of complex backgrounds, the fused multi-level features are dissected on the channel, salient regions are individually excavated, and features originating from different channels are adaptively amalgamated.
The AMANet is introduced by embedding the AMAM between the backbone network and the FPN, demonstrating its versatility as it can be readily inserted between different frameworks to improve object detection.
Extensive experiments on two large-scale SAR ship detection datasets validate the effectiveness of our proposed AMANet method, showing its superiority over state-of-the-art methods. 
\textbf{In the Future.} Although AMANet has demonstrated its effectiveness on two large-scale SAR datasets and multiple detection frameworks, these are all based on CNN architecture backbone networks. We plan to explore the effectiveness of AMANet under the Transformer backbone network further.

\bibliographystyle{IEEEtran}
\bibliography{ref} 

\begin{IEEEbiography}[{\includegraphics[width=1in,height=1.25in,clip,keepaspectratio]{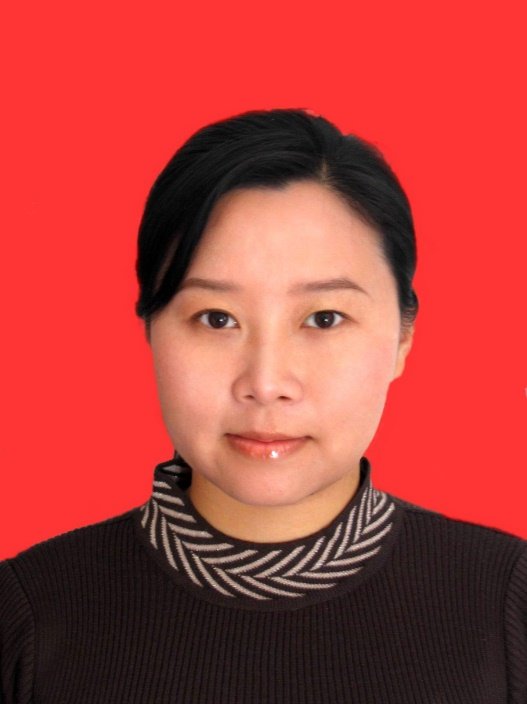}}]{XIAOLIN MA} received the M.S. degree from Communication University of China, Beijing, China. She is currently working with Army Engineering University, 
Shijiazhuang. Her main research interests include object detection and signal processing.
\end{IEEEbiography}

\begin{IEEEbiography}[{\includegraphics[width=1in,height=1.25in,clip,keepaspectratio]{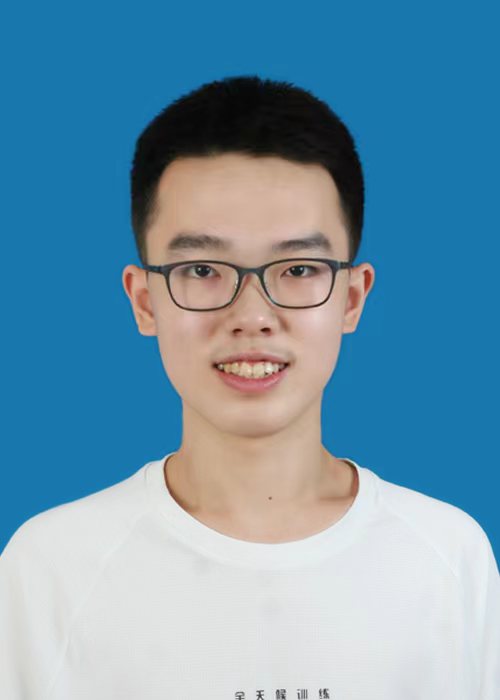}}]{JUNKAI CHENG} is currently an undergraduate student majoring in Automatic Control at Hebei University of Technology, Shijiazhuang, Hebei, China. His research interests include target recognition.
\end{IEEEbiography}

\begin{IEEEbiography}[{\includegraphics[width=1in,height=1.25in,clip,keepaspectratio]{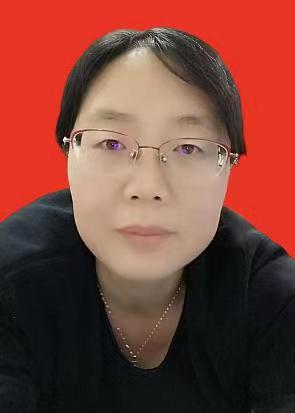}}]{AIHUA LI} {AIHUA LI} received the M.S. degree from Hebei University Of Science and Technology, Shijiazhuang,China. She is currently working with Army Engineering University, 
Shijiazhuang. Her main research interests include object detection.
\end{IEEEbiography}

\begin{IEEEbiography}[{\includegraphics[width=1in,height=1.25in,clip,keepaspectratio]{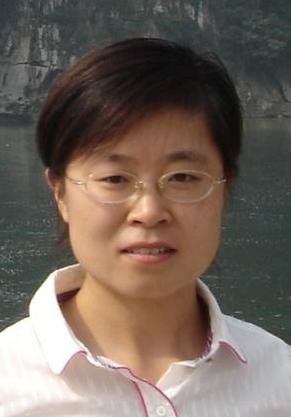}}]{YUHUA ZHANG} received the  Ph.D. degrees from the National University of Defense Technology, Changsha, China. She is currently working in Army Engineering University, Shijiazhuang. Her main research interest includes remote sensing image processing.
\end{IEEEbiography}

\begin{IEEEbiography}[{\includegraphics[width=1in,height=1.25in,clip,keepaspectratio]{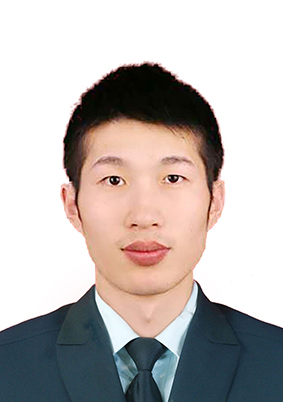}}]{ZHILONG LIN}  received the B.S. and M.S. degrees from Army Engineering University,  Shijiazhuang, China. He is currently an lecturer with the Department of Unmanned Aerial Vehicle Engineering, Army Engineering University, Shijiazhuang. His current research interests include object detection and visual tracking.
\end{IEEEbiography}

\EOD

\end{document}